\documentclass[10pt, a4paper]{article}
\usepackage{tabularx}
\usepackage{makecell}
\usepackage{array}
\usepackage{listings}
\usepackage{multirow}

\usepackage{booktabs}    
\usepackage{tabularx}    
\usepackage{makecell}    
\usepackage{caption}     
\usepackage{subcaption}

\usepackage{lrec2026} 
\usepackage{booktabs}

\title{MM-Conv: A Multimodal Dataset and Benchmark for Context-Aware Grounding in 3D Dialogue}

\name{\parbox{0.85\textwidth}{\centering Anna Deichler, Jim O'Regan, Fethiye Irmak Dogan, Lubos Marcinek, Anna Klezovich, Iolanda Leite, and Jonas Beskow}}
\address{KTH Royal Institute of Technology, 
         Stockholm, Sweden\\
         \{deichler, joregan, fidogan, lubosm, annkle, iolanda, beskow\}@kth.se\\
        }

\abstract{
Grounding language in the physical world requires AI systems to interpret references that emerge dynamically during conversation. While current vision-language models (VLMs) excel at static image tasks, they struggle to resolve ambiguous expressions in spontaneous, multi-turn dialogue. We address this gap by introducing (1) a benchmark for referential communication in dynamic 3D environments, built from 6.7 hours of egocentric VR interaction with synchronized speech, motion, gaze, and 3D scene geometry, and (2) a two-stage grounding pipeline that explicitly resolves conversational ambiguity before visual localization. The benchmark includes over 4,200 manually verified referring expressions spanning full, partitive, and pronominal types. Our contextual rewriting approach improves grounding performance by 11–22 percentage points on average, with a pure detector (GroundingDINO) reaching 56.7\% on pronominals after rewriting—nearly double the best end-to-end baseline.  Results demonstrate that decoupling linguistic reasoning from visual perception is more effective than end-to-end approaches for conversational grounding.
\\ \newline
}

\begin{document}

\maketitleabstract

\section{Introduction}

Understanding and resolving referring expressions in situated, real-world contexts remains a core challenge for multimodal AI. While recent progress in vision-language models (VLMs) has brought significant advances in grounding natural language to images and videos, these models often fall short when it comes to reference resolution in dynamic, embodied environments. Here, language is rarely disambiguated by words alone; instead, meaning is distributed across multiple modalities, including speech, gesture, gaze, and spatial context, unfolding over time and grounded in interaction~\cite{10.1145/3136755.3136795}. These capabilities are critical for applications such as home-assistive robots or AR-based guides, where agents must interpret ambiguous phrases like ‘that one’ in context.

Existing benchmarks for referring expression grounding, such as ScanRefer~\cite{chen2020scanrefer} or ReferIt3D~\cite{achlioptas2020referit3d}, have driven progress but are limited by their reliance on scripted, single-turn textual descriptions and static 3D scenes.  This leaves a critical gap in evaluating models on the spontaneous, messy, and multimodal nature of real human dialogue.   
To address this, we introduce a new benchmark that shifts the focus to the challenge of resolving spontaneous, conversational references as they emerge naturally in rich, situated dialogues. These expressions, such as ``this one here'', ``the yellowish thing'', or simply ``it'', are inherently ambiguous unless grounded in a shared perceptual and interactional context.

In realistic, situated dialogue, language is often insufficient on its own. The problem of referential understanding has been approached using descriptive features in language~\cite{shridhar2020ingress, 9889368, dogan2024semantically}, but in spontaneous interaction, non-verbal cues are crucial for resolving ambiguity, with studies showing that referential gaze is tightly synchronized with speech \cite{STAUDTE2011268} and that gestures provide effective guidance for disambiguating objects \cite{zhang2024gaze, 10.1145/2559636.2559657}.  Recent work has explored learning to generate such pointing behaviors for embodied agents  \cite{deichler2023pointing}.

Early benchmarks miss core conversational phenomena—ellipsis, anaphora, and multimodal deixis, common in situated dialogue. Current VLMs, largely trained on text–image pairs, struggle to resolve such ambiguities without nonverbal cues (e.g., gaze, gesture) and often lack explicit, structured reasoning for highly underspecified references. Our contribution is the first benchmark that \emph{synchronizes} these signals with conversational language, setting the stage for truly multimodal grounding.

To support this, we built a 6.7-hour dataset of immersive, multimodal interactions with synchronized speech, motion, gaze,  and 3D scene geometry. We annotated 4{,}211 naturally occurring referring expressions and designed an evaluation framework spanning explicit nouns to highly contextual pronouns. Results reveal critical limitations of state-of-the-art VLMs for temporally grounded references and demonstrate the effectiveness of a modular, ambiguity-first pipeline for context-aware grounding.

Our main contributions are:
\begin{itemize}
\item \textbf{A new benchmark for spontaneous, multimodal referential grounding}, with 4,211 annotated expressions from 6.7 hours of conversational data in dynamic 3D environments.
\item \textbf{A two-stage grounding pipeline} that separates conversational disambiguation from visual localization, improving performance by 11--22 percentage points across reference types.
\item \textbf{Systematic evaluation} revealing that (1) current VLMs struggle with conversational references despite native context support, and (2) modular architectures outperform end-to-end approaches for situated grounding.
\item \textbf{A rich, reusable dataset} with synchronized full-body motion, gaze, facial expression, and egocentric video to support downstream research in embodied interaction.
\end{itemize}


The rest of the paper is structured as follows: Section 2 describes related work in referential understanding and grounding in vision-language models. Section 3 describes the dataset and the annotations, and Section 4 presents the human and VLM experiments.   

\section{Related Work}

\subsection{Referential Understanding Benchmarks}

\begin{table*}[t]
\small
\caption{Comparison with prior multimodal referential benchmarks.}
\begin{tabular}{lllll}
\toprule
\textbf{Dataset} & \textbf{Modalities} & \textbf{Scale} & \textbf{Embodied} & \textbf{Dialogue} \\
\midrule
ScanRefer (2020) & Text, 3D & 51.6k, 800 scenes & No & One-shot \\
YouRefIt (2021) & Text, RGB, Gesture & 4k, 432 scenes & Yes & Multi-turn \\
TEACh (2022) & Text, Sim3D, Actions & 3k dialogues & Nav/Manip & Multi-turn \\
\textbf{Ours} & \textbf{Speech, RGB, Depth, Motion, Gaze} & \textbf{4.2k, 5 scenes} & \textbf{Full VR} & \textbf{Spontaneous} \\
\bottomrule
 \label{tab:related}
\end{tabular}
\end{table*}

Foundational datasets like ScanRefer~\cite{chen2020scanrefer} and ReferIt3D~\cite{achlioptas2020referit3d} established 3D referring expression grounding but rely on single-turn, text-only descriptions. While YouRefIt~\cite{chen2021yourefit} added multi-turn dialogue with gesture, it uses third-person video. TEACh~\cite{padmakumar2022teach} includes embodied dialogue but lacks continuous multimodal streams (gaze, motion). 

Our work addresses this gap by integrating temporally synchronized speech, full-body motion, gaze, and egocentric vision from immersive VR, enabling the investigation of reference resolution with all modalities naturally co-occurring in interaction. Table~\ref{tab:related} summarizes key differences.

\subsection{Vision–Language Models for Grounding}
Vision-language models (VLMs) for grounding have evolved significantly, moving from dual-encoder architectures to more flexible generative models. This progression is critical for handling the spontaneous, conversational language, which stands in contrast to the clean, template-like queries found in traditional referring expression benchmarks. Models like GroundingDINO \cite{liu2024grounding}, built on CLIP \cite{radford2021learning} embeddings, perform well on short to medium-length prompts; however, their ability to generalize to longer, more conversational inputs is limited. In contrast, Florence-2 \cite{xiao2024florence} integrates a co-trained causal language model that supports prompt-driven decoding over both visual and textual inputs. This makes Florence-2 particularly well-suited for grounding tasks in ecologically valid settings where referring expressions are embedded in natural dialogue. Recent advances in GPT-style vision-language models, such as GroundingGPT \cite{li2024groundinggpt}, further push this boundary by leveraging large-scale, decoder-only architectures that unify visual and textual modalities in a generative framework. Table~\ref{tab:models} provides a comprehensive comparison of these vision-language grounding models, highlighting their architectural differences and capabilities.

\section{Dataset}

\begin{table*}[t]
\small
\centering
\setlength{\tabcolsep}{3pt}
\renewcommand{\arraystretch}{1.2}
\caption{Comparison of vision-language grounding models, including text/visual encoders, training status, and notable capabilities.}
\begin{tabularx}{\textwidth}{@{}p{1.9cm} p{1.5cm} p{2.5cm} p{1.5cm}  c X@{}}

\toprule
\textbf{Model} & \textbf{Text Enc.} & \textbf{Visual Enc.} & \textbf{Visual In.} & \textbf{Token}  & \textbf{Notable Features} \\
\midrule
Grounding DINO \cite{liu2024grounding} & BERT & DETR (Swin) & RGB (2D) & 256 & Referring expression grounding via cross-attention; zero-shot localization with box output. \\
Florence-2 \cite{xiao2024florence} & Transformer (BART) & DaViT & RGB (2D) & 1024 & Foundation model with prompt-driven generation; supports grounding using polygon or box output. \\
KOSMOS-2 \cite{peng2023kosmos2} & Transformer (causal LM) & ViT (patch14-224) & RGB (2D) & 2048  & Grounds text spans to boxes via ``location tokens''; outputs grounded captions/VQA with box coordinates (open-vocab grounding). \\
Ferret \cite{you2024ferret} & Vicuna-style LLM & CLIP ViT-L/14 (region features) & RGB (2D) & 4096  & Referring and region-level understanding via explicit region features; supports referring/grounded dialogue with box-based region specification. \\
GroundingGPT \cite{li2024groundinggpt} & Vicuna-v1.5 & CLIP ViT-L/14, Q-former (video) & RGB (2D), Video & 4096  & Multi-modal model achieving SOTA performance in image, video, and audio grounding. \\
Qwen2.5-VL
(Bai, 2025)  &  Qwen2.5
(causal LM)  &  ViT (window attn, SwiGLU)  &  RGB (2D),
Video  &  4--16k
(dyn.)  &  Dynamic resolution with mRoPE; native
bounding-box grounding; supports multi-turn
dialogue and visual agent tasks.  \\

\bottomrule
\end{tabularx}
\label{tab:models}
\end{table*}

To create a benchmark for spontaneous, multimodal referential grounding, we collected a new dataset consisting of 6.7 hours of interaction data recorded during a referential communication task in a virtual environment. Our primary goal was to capture the richness of embodied dialogue, including synchronized speech, full-body motion, gaze, and 3D scene geometry, providing a foundation for studying grounding in a controlled yet naturalistic setting. Further details on data collection can be found in the Appendix.

\subsection{Dataset Collection and Experimental Setup}
The dataset was collected in dyads following a typical instruction-giver/follower paradigm, common in cognitive studies of referential communication. The main actor (instruction-giver), equipped with a full-body motion capture suit, finger-tracking gloves, and a VR headset with gaze and facial tracking, described objects and spatial arrangements within simulated apartment environments (AI2-THOR~\cite{kolve2017ai2}). The interlocutor (instruction-follower) responded naturally but was instructed not to introduce new referents, to elicit a high density of spontaneous referring expressions and associated nonverbal behaviors from the main actor. We selected five apartment environments from AI2-THOR, each containing a diverse set of interactable objects, with 2–3 short scenarios per room to vary discourse goals (e.g., showing a new apartment, landlord inspection, interior-designer suggestions) while maintaining ecological validity. This setup yields synchronized speech, full-body motion, gaze, facial expressions, and structured 3D scene graphs with object-level metadata for every frame. Further details on the data collection and annotation processes are provided in the Appendix.

\paragraph{Main–Interlocutor protocol.}
\begin{itemize}
\item \textbf{Roles}: Main actor introduces and describes objects; interlocutor reacts but avoids adding novel referents, preserving a single discourse locus per segment.
\item \textbf{Scenarios}: Per room, 2–3 role-play scenarios (e.g., “bragging about a new apartment,” “landlord inspection,” “interior designer tips”) encourage varied yet natural reference types (full, partitive/attribute, pronominal).
\end{itemize}

\paragraph{Hardware and synchronization.}
Collection used OptiTrack mocap (50 markers), MANUS MetaGloves (finger tracking), and Meta Quest Pro (binocular gaze, 52 facial blendshapes). All streams were synchronized via SMPTE timecode injected into Unity, with headset pose calibrated to mocap for egocentric alignment.


\paragraph{Environments and objects.}
Five apartment rooms from AI2-THOR (Unity) were pre-selected to span common household categories and spatial layouts; scene graphs (per frame) were exported from the simulator with canonical object identifiers to support instance-level grounding. (See Appendix for object distributions.
\begin{figure}[ht]
  \centering
  \includegraphics[width=0.98\linewidth]{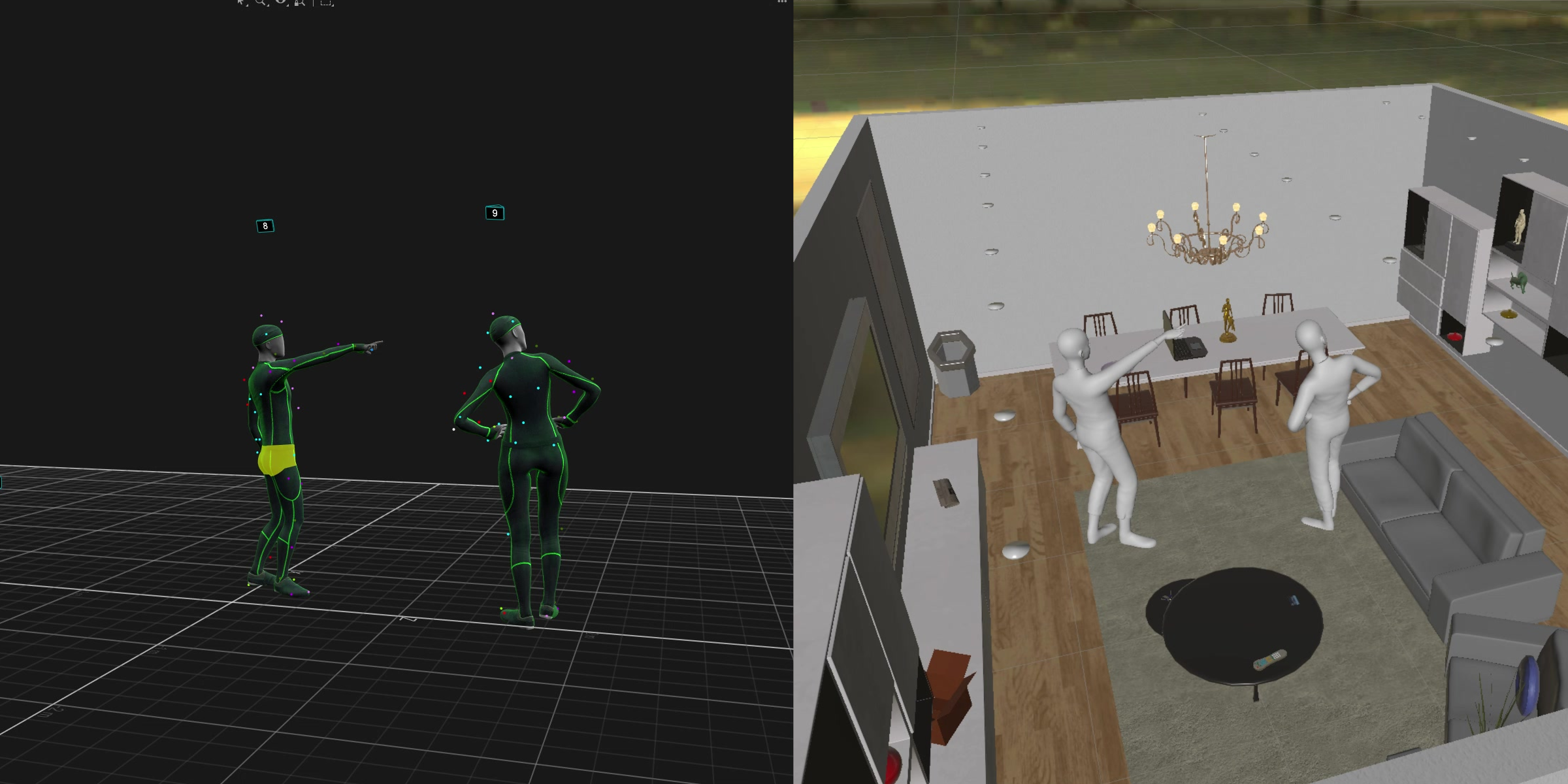}
  \caption{The multimodal data collection environment. A participant in a full-body motion capture suit and VR headset (left) interacts within a virtual scene. Their actions are rendered in real-time in the AI2-THOR simulator (right), enabling the synchronized capture of egocentric vision, speech, motion, and 3D scene geometry.}
  \label{fig:crowdsourcing_image}
\end{figure}

\subsection{Annotation}
All modalities were aligned using the shared SMPTE timecode recorded during capture (audio-mocap-gaze/face-simulation). The timecode stream was also recorded inside the simulator to ensure frame-accurate association between referring expressions and rendered egocentric frames. This synchronization enables precise mapping from word-level timestamps to simulator frames for RGB, depth, and per-pixel instance masks.

To transform the raw data into a curated benchmark, we developed a multi-stage annotation pipeline.

\subsubsection{Speech Transcription}

The audio data was transcribed using WhisperX~\cite{bain23_interspeech}, which was prompted with a manually transcribed portion of the audio to encourage the inclusion of filled pauses, repeated words, and other aspects of spontaneous speech. The resulting transcripts were then edited by human annotators to ensure precision. 
The corrected transcripts were then aligned using the CTC-forced-aligner to obtain word-level timings. Further information on speech transcription can be found in the Appendix \footnote{For word-level alignment, we used a CTC forced aligner: \url{https://github.com/MahmoudAshraf97/ctc-forced-aligner}.}

\subsubsection{Reference Annotation}
We developed a two-stage pipeline using GPT-4o to generate visually grounded reference annotations. First, GPT-4o produced topic annotations for each utterance—based on WhisperX VAD segmentation and the visible objects in the Unity scene—highlighting discourse focus and shifts in attention. Next, GPT-4o parsed the utterances to classify referring expressions into three categories: full noun phrase, partitive/attribute noun phrase, and pronominal.  All GPT-4o classifications were manually verified and corrected where necessary by the research team, with most corrections involving boundary cases between partitive and pronominal types.%

\paragraph{Categories.}\mbox{}\\

\textit{Full noun phrases (full NP)} are explicit, uniquely identifying descriptions (e.g., “the black sofa in the corner”). 

\textit{Partitive/attribute noun phrases (partitive NP)}, used here as a cover term for underspecified references, refer to parts, features, subsets, salient attributes, or spatial/deictic indicators of an object (e.g., "the cloud (in the painting)", "the rubber thing (part of the lamp)", "the yellowish one", "there") that are insufficiently specific to uniquely identify a referent without additional context. We subsume spatial indicators under this category rather than introducing a separate type, as they share this defining property of underspecification.

\textit{Pronominal references (pronouns)} are expressions like “it”, “that”, or “those”, which rely heavily on discourse context and shared perceptual attention.





These categories align with established distinctions in referring expression research: full NPs as definite descriptions \cite{dale1995computational,kazemzadeh2014referitgame}, partitive/attribute NPs for part-whole and property-based reference \cite{vandersluis2001generating,viethen2008spatial}, and pronominal references for deictic or anaphoric pronouns \cite{hobbs1978resolving,STAUDTE2011268}. Together, they span a continuum from lexically explicit to contextually dependent reference.

\paragraph{Grounding and validation.}
We grounded each expression to specific scene objects using raycasting and Unity-derived object masks. All links between referring phrases and referents were manually verified for correctness, yielding rich, context-sensitive annotations that combine linguistic and perceptual cues.

\subsection{Final Dataset Format}

To build a usable multimodal dataset for grounded language modeling, we aligned the referring expressions with the visual stream via word-level timestamps, yielding the exact frame where each reference occurs. For that frame, we extract synchronized egocentric assets: (1) an RGB image, (2) a per-pixel metric depth map, and (3) a segmentation mask with per-pixel object IDs. Frames are rendered in Blender for high-quality egocentric imagery, and the pipeline’s flexible camera setup supports alternative viewports (e.g., third-person views) for future tasks. This format grounds language in both semantics and spatial structure, supporting reference resolution, object localization, and multimodal grounding in 3D settings.



\begin{figure}[ht]
  \centering

  \includegraphics[width=0.48\linewidth]{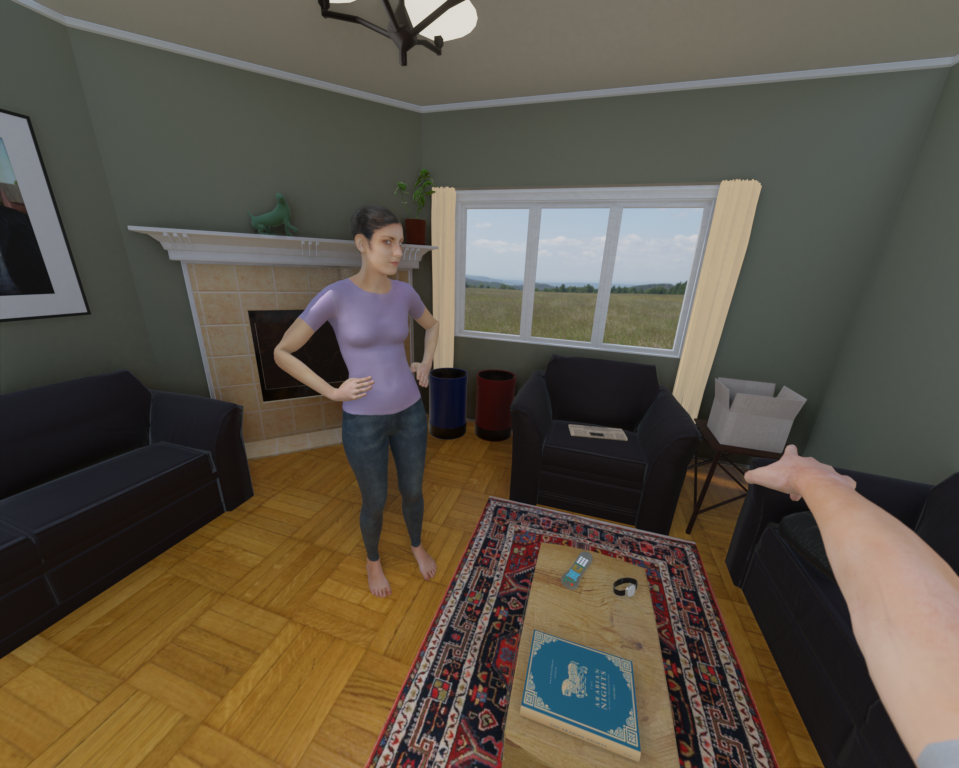}
  \caption*{(a) Visual egocentric RGB scene rendered from the main speaker viewpoint for reference `box'.}
  
  \vspace{0.5em}
  \begin{minipage}{0.48\linewidth}
    \centering
    \includegraphics[width=\linewidth]{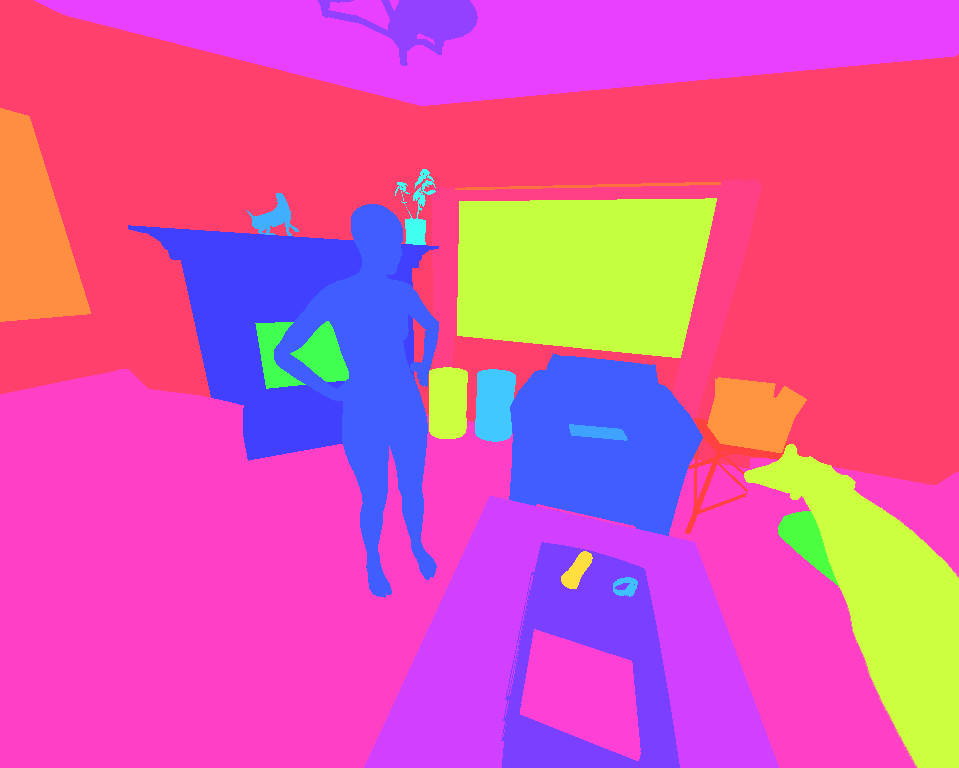}
    \caption*{(b) Segmentation mask.}
  \end{minipage}%
  \hfill
  \begin{minipage}{0.48\linewidth}
    \centering
    \includegraphics[width=\linewidth]{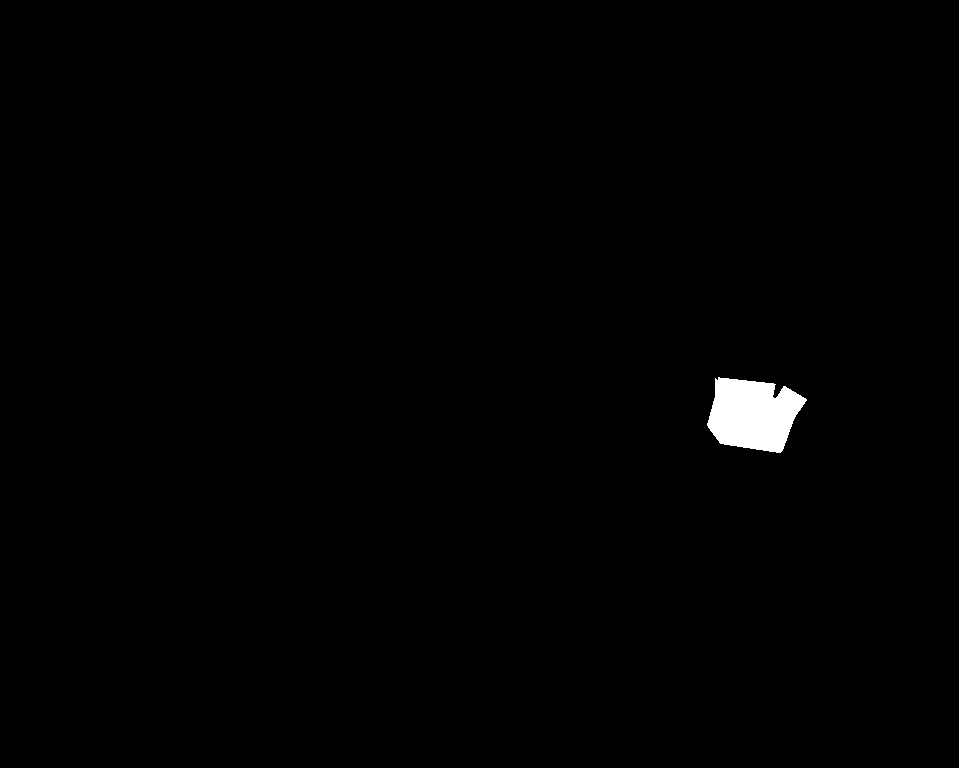}
    \caption*{(c) Target object.}
  \end{minipage}
  

  

  \caption{A grounded data sample from the benchmark. For a referring expression like "box", our dataset provides synchronized data streams: (a) The egocentric RGB view with the ground-truth referent (box) highlighted by a green segmentation mask. (b) The raw RGB image. (c) The corresponding depth map. This structure facilitates evaluation on precise, pixel-level grounding.}
  \label{fig:triple_layout}
\end{figure}

\subsection{Dataset statistics}

Our dataset comprises a total of 4,211 referring expressions collected over 6.7 hours of recorded interaction. Each expression was annotated for linguistic form, visibility, and grounding success. Overall, virtually all referring expressions were successfully matched to objects in the environment, demonstrating the high referential clarity of the collected data. After filtering for expressions directed at objects not visible at the moment of reference, a final set of 4,001 references was established for analysis. The data also reveals natural variations in referential strategies across different participants and recording environments (Tables~\ref{tab:agg_pid_summary} and \ref{tab:agg_room_summary}), reflecting diverse interaction styles and the trade-offs between scene complexity and referential clarity.

\begin{table}[h]
\small
\centering
\caption{Aggregate and participant-wise reference statistics for main speaker (m) and interlocutor (i).}
\label{tab:agg_pid_summary}
\setlength{\tabcolsep}{4pt}
\begin{tabular}{lrrrr}
\toprule
\textbf{pid} & \textbf{\# files} & \textbf{Total Refs (m)} & \textbf{Total Refs (i)} \\
\midrule
All & 54 & 4211 & 1,547 \\
3   & 8  & 432  & 220  \\
4   & 13 & 974  & 296  \\
5   & 12 & 925  & 391  \\
6   & 10 & 1008 & 322  \\
7   & 11 & 872  & 318  \\
\bottomrule
\end{tabular}
\end{table}

\begin{table}[h]
\small
\centering
\caption{Aggregate and room-wise reference statistics}
\label{tab:agg_room_summary}
\setlength{\tabcolsep}{4pt}
\begin{tabular}{lrrr}
\toprule
\textbf{Room} & \textbf{Total Refs} & \textbf{Invisible} & \begin{tabular}[c]{@{}r@{}}\textbf{Unique Objs}\\\textbf{(avg)}\end{tabular} \\
\midrule
All  & 4211 & 210 & 23.74 \\
209  & 807  & 18  & 20.42 \\
210  & 942  & 51  & 22.75 \\
211  & 726  & 33  & 23.22 \\
222  & 844  & 37  & 30.38 \\
227  & 892  & 71  & 24.00 \\
\bottomrule
\end{tabular}
\end{table}

A key finding is the prevalence of context-dependent language, which is often underrepresented in existing multimodal datasets. As shown in Table~\ref{tab:ref_distribution}, pronominal references (e.g., “it,” “that one”) are especially common. In the main (egocentric) view, pronouns account for 2,078 expressions (49.3\% of 4,211 total; 1,591 single + 487 multiple). In the inter (partner) view, pronouns are even more dominant, totaling 1,070 expressions (69.2\% of 1,547; 901 single + 169 multiple). This high frequency underscores the central role of pronouns in natural dialogue and highlights a core challenge for current models, which often struggle with such context-dependent forms. By explicitly including and annotating these expressions, our dataset fills a critical gap and enables the development of grounded models that better capture the fluid, context-dependent nature of situated language. The categorization of expression types was obtained using GPT-4o~\cite{openai2024gpt4o}, guided by a custom prompt (described in the Appendix) and manually verified for correctness.

Beyond overall rates, the two perspectives reflect distinct discourse roles. Main speakers more often introduced objects into the conversation, yielding relatively more full mentions, whereas the interlocutor tended to react to already-salient items, relying heavily on pronominal and deictic forms to maintain common ground (Table~\ref{tab:ref_distribution}).

\begin{table}[h]
\small
\caption{Reference distribution by type and interlocutor (single/multiple).}
\label{tab:ref_distribution}
\begin{tabular}{lrrr|rrr}
\toprule
& \multicolumn{3}{c}{\textbf{Main}} & \multicolumn{3}{c}{\textbf{Inter}} \\
\cmidrule(lr){2-4} \cmidrule(lr){5-7}
& full & part & pron & full & part & pron \\
\midrule
Single & 1215 & 435 & 1591 & 248 & 48 & 901 \\
Multiple & 348 & 135 & 487 & 137 & 44 & 169 \\
\bottomrule
\end{tabular}
\end{table}

\section{Experiments}
To gain a comprehensive understanding of referential grounding, we adopt a dual evaluation strategy combining crowd-sourced human judgments and vision-language model (VLM) evaluation. Human studies serve not only to benchmark model performance but also to validate the interpretability of our dataset. However, it is important to note that crowd-sourcing takes place outside the original interaction context. As such, crucial 3D spatial relationships, scene dynamics, and nonverbal cues available to the original speaker and listener may be absent. Complementing this, we evaluate state-of-the-art VLMs (GroundingGPT, Ferret, Kosmos-2, Florence-2, and Qwen2.5-VL) to understand how well models generalize to open-ended, conversational referring expressions. Together, these evaluations shed light on the challenges of reference resolution in both human and computational agents.
\subsection{Experimental Setup}
We split our study into subsets based on the referential expression categories in Table \ref{tab:ref_distribution}. We focus on single-object references, which we further divide into 3 subsets, based on the referring expression categories. 

\begin{enumerate}
\item \textbf{Exact noun phrases:} Explicit object names resembling classic referring expression benchmarks, where grounding depends on direct lexical match.
\item \textbf{Filtered partitives:} Partitive NPs with abstract or spatial terms (e.g., “the area,” “there”) removed, retaining only those grounded in identifiable scene elements.
\item \textbf{Subsampled pronominals:} A representative subset of pronouns requiring discourse or visual context for correct resolution.
\end{enumerate}

\paragraph{Evaluation Metrics.}
We evaluate visual grounding using Intersection over Union (IoU) between predicted and ground-truth bounding boxes, with ground-truth boxes derived from instance segmentation masks.
We report accuracy at two standard thresholds, Acc@0.3 and Acc@0.5, as well as mean IoU across samples.




\subsection{Human Evaluation}

\begin{figure}[ht]
  \centering
  \includegraphics[width=0.85\linewidth]{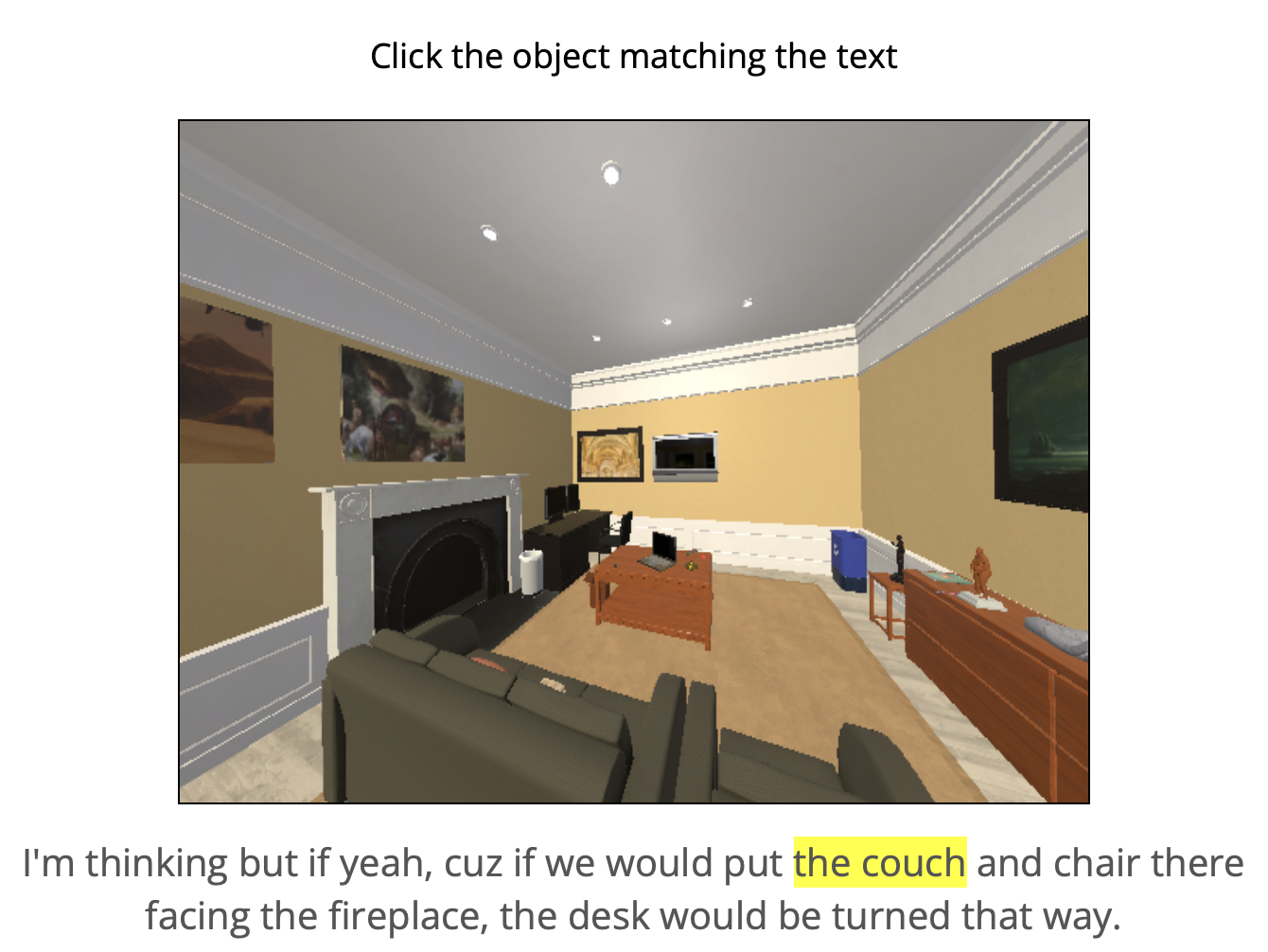}
  \caption{The interface for our human evaluation study. Crowd-workers were presented with an egocentric image and a corresponding utterance from the dataset. They were tasked with clicking on the object being referred to, providing a human baseline for reference resolution.}
  \label{fig:crowdsourcing_image1}
\end{figure}

Human evaluations were conducted using the crowd-sourcing platforms Prolific \cite{prolific2024} and Cognition.run \cite{cognitionrun2024}. Participants were presented with the first-person-view images and corresponding utterance with highlighted referring expressions, and were asked to click on the referred object in the picture (see Figure \ref{fig:crowdsourcing_image1}). We tracked the use of exact noun phrases, partitive noun phrases, and pronoun-based references under the single object condition, while also comparing isolated utterances with context-based utterances. In order to add context, we sampled the previous five utterances, selecting only those that matched the topic; in the event that this did not yield any text, we selected text from the word-level transcriptions for the previous 20 seconds. This contextual history was then added to the utterance text. Attention checks were included to ensure data quality.


Each of the 1940 stimuli was evaluated by three participants. The total number of participants included in the analysis, after discarding those 4.9\% participants who failed attention checks, is 78.

The percentage of clicks that are within the bounding box around the ground-truth image mask was recorded as accurate. A three-way majority agreement was also calculated for each of the data subsets.

\begin{table}[ht]
\centering
\caption{Crowd-sourcing results for different types of referring expressions.}
\label{tab:crowd_sourcing}
\setlength{\tabcolsep}{5pt}
\begin{tabular}{lccc}
\toprule
\textbf{Subset} & \textbf{Acc} & \begin{tabular}[c]{@{}c@{}}Majority\\Agreement\end{tabular} & \begin{tabular}[c]{@{}c@{}}Median\\time (m)\end{tabular} \\
\midrule
full np, w ctx  & 62.45\% & 60.92\% & 18:09 \\
full np, no ctx    & 73.18\% & 74.43\% & 16:19 \\
part np, w ctx  & 60.99\% & 61.96\% & 28:09 \\
part np, no ctx    & 47.93\% & 47.01\% & 15:56 \\
pron, w ctx     & 55.42\% & 55.16\% & 25:12 \\
pron, no ctx       & 37.43\% & 33.33\% & 18:28 \\
\bottomrule
\end{tabular}
\end{table}

The crowd-sourced evaluation (Table~\ref{tab:crowd_sourcing}) reveals clear trends across expression types and context conditions. Full NPs without context achieved the highest accuracy (73.18\%) and agreement (74.43\%), indicating that explicit references are easiest to resolve. Adding context slightly reduced both (62.45\%, 60.92\%), suggesting that additional information can introduce ambiguity. The smaller difference in median completion time between with- and without-context conditions further implies participants relied less on context when full NPs were explicit.
Context substantially benefits underspecified forms. Partitive NPs improved from 48\% to 61\% accuracy, while pronominals rose from 37\% to 55\%. Humans achieved non-trivial accuracy on pronominals even without linguistic context (37\%), likely by leveraging visual cues such as the speaker's pointing gesture visible in the egocentric view. In contrast, VLMs often grounded to the hand itself or nearby salient regions rather than following the deictic gesture to the intended referent.



\paragraph{Inter-Annotator Reliability.}
We assessed annotation consistency using object-level agreement with a 30px clustering tolerance.
Annotators achieved 75--83\% unanimous agreement across conditions (Krippendorff’s $\alpha$ = 0.42--0.66),
with tight localization (24--44px) when identifying the ground-truth referent.
Unanimous misses—where all annotators clicked outside the target mask—fell into three patterns:
near-misses due to conservative mask boundaries (15--20\%),
coherent agreement on alternative referents reflecting genuine ambiguity ($\sim$30\%),
and scattered responses indicating true confusion (24--51\%, highest for pronouns without context).
The median pairwise distance among misses (11--64px) remains far below chance ($\sim$270px),
indicating consistent interpretations even when diverging from ground truth.

\subsection{VLM evaluation}

To understand how vision-language models handle natural language grounding in situated interaction, we evaluated several state-of-the-art approaches on our benchmark. We compared end-to-end VLMs against a novel two-stage pipeline designed to explicitly handle conversational context through linguistic disambiguation prior to visual grounding.

\begin{table*}[h!]
\small
\centering
\caption{VLM Grounding Performance. Match Rate at IoU 0.5/0.3 (M@.5/M@.3) and mean IoU for matched samples (mIoU@.5). Best M@.5 per row in \textbf{bold}.}
\label{tab:vlm_results_single}
\setlength{\tabcolsep}{4pt}
\begin{tabular}{llccccc}
\toprule
\textbf{Expr. Type} & \textbf{Context} & \textbf{GroundingGPT} & \textbf{Ferret} & \textbf{Kosmos-2} & \textbf{Florence-2} & \textbf{Qwen2.5-VL} \\
\midrule
\multirow{2}{*}{Full NP}
& w/o ctx & 37.9/47.3 & 39.1/45.2 & 42.6/48.1 & 46.3/52.6 & \textbf{51.8}/60.9 \\
& w/ ctx  & 28.2/41.1 & 40.9/47.9 & 42.0/48.2 & 28.0/33.1 & \textbf{53.2}/62.0 \\
\midrule
\multirow{2}{*}{Partitive NP}
& w/o ctx & 14.2/16.3 & 14.7/16.7 & 16.5/17.9 & 16.3/18.8 & \textbf{22.5}/26.4 \\
& w/ ctx  & 14.7/20.9 & 18.3/21.6 & 19.7/22.2 & 21.6/23.6 & \textbf{29.6}/34.2 \\
\midrule
\multirow{2}{*}{Pronominal}
& w/o ctx & 4.7/6.9 & 5.1/7.4 & 6.3/9.1 & \textbf{9.2}/10.1 & 5.9/8.3 \\
& w/ ctx  & 10.7/16.0 & 12.3/15.4 & 9.3/11.4 & 25.8/28.5 & \textbf{30.4}/38.0 \\
\bottomrule
\end{tabular}
\end{table*}

\begin{table*}[t]
\small
\centering
\caption{Rewrite-based Grounding Results. Referring expressions are rewritten by Qwen2.5-VL then grounded. Match Rate at IoU 0.5 (M@.5) shown with improvement over baseline in parentheses.}
\label{tab:rewrite_results}
\setlength{\tabcolsep}{6pt}
\begin{tabular}{lcccccc}
\toprule
& \multicolumn{2}{c}{\textbf{Full NP}} & \multicolumn{2}{c}{\textbf{Partitive NP}} & \multicolumn{2}{c}{\textbf{Pronominal}} \\
\cmidrule(lr){2-3} \cmidrule(lr){4-5} \cmidrule(lr){6-7}
\textbf{Model} & Baseline & Rewrite & Baseline & Rewrite & Baseline & Rewrite \\
\midrule
Qwen2.5-VL    & 53.2 & 54.4 \textcolor{blue}{(+1.2)}  & 29.6 & 40.8 \textcolor{blue}{(+11.2)} & 30.4 & 50.3 \textcolor{blue}{(+19.9)} \\
Florence-2    & 28.0 & 49.1 \textcolor{blue}{(+21.1)} & 21.6 & 39.7 \textcolor{blue}{(+18.1)} & 25.8 & 48.9 \textcolor{blue}{(+23.1)} \\
GroundingDINO & —\textsuperscript{*} & \textbf{61.1} & —\textsuperscript{*} & \textbf{49.5} & —\textsuperscript{*} & \textbf{56.7} \\
\midrule
Avg. Gain\textsuperscript{†} & \multicolumn{2}{c}{+11.2} & \multicolumn{2}{c}{+14.7} & \multicolumn{2}{c}{+21.5} \\
\bottomrule
\multicolumn{7}{l}{\textsuperscript{*}No conversational capability; baseline not applicable.} \\
\multicolumn{7}{l}{\textsuperscript{†}Average gain for Qwen2.5-VL and Florence-2 only.} \\
\end{tabular}
\end{table*}

\subsubsection{Baseline VLM Performance}

Table~\ref{tab:vlm_results_single} presents baseline results for five VLMs: GroundingGPT~\cite{li2024groundinggpt}, Ferret \cite{you2024ferret}, Kosmos-2 \cite{peng2023kosmos2}, Florence-2~\cite{xiao2024florence}, and Qwen2.5-VL \cite{bai2025qwen25vl}. We evaluated each model in two conditions: without context (only the referring utterance) and with context (previous 5 topic-matched utterances or 20 seconds of dialogue).

Full NPs achieve moderate success, with Qwen2.5-VL reaching 53.2\% with context. Context provides minimal benefit (+2 points), as these expressions are already self-contained. Partitive NPs prove challenging for all models (14--30\% range). Qwen2.5-VL achieves 29.6\% with context, but this remains far below human performance (61\% in our crowd-study). Models struggle to infer part-whole relationships from conversational cues alone.

Pronominal references reveal the starkest limitation. Without context, all models fail catastrophically (5--9\%), as expected for expressions like ``it'' or ``that.'' With context, performance improves to 9--30\%, but remains poor. Even Qwen2.5-VL, which can explicitly process dialogue history, achieves only 30.4\% on pronominals—indicating that simply providing context to VLMs is insufficient.

\subsubsection{Two-Stage Grounding Pipeline}

The baseline results expose a critical gap: VLMs struggle to leverage conversational context for ambiguity resolution. We hypothesize this stems from conflating two distinct challenges—linguistic disambiguation and visual localization—in a single forward pass. To address this, we propose a modular pipeline.

In Stage 1 (Contextual Rewriting), Qwen2.5-VL rewrites ambiguous references into explicit descriptions (e.g., ``it'' $\rightarrow$ ``the wall painting'', ``the figures'' $\rightarrow$ ``the figures on the rug''). The rewriter resolves pronouns using dialogue history, adds part-whole relationships for underspecified expressions, and preserves disambiguating qualifiers. Importantly, the rewriter sees only text—no images—making this pure linguistic reasoning.

In Stage 2 (Visual Grounding), the disambiguated expression is passed to a visual grounding model (Florence-2, GroundingDINO, or Qwen2.5-VL) without fine-tuning. Critically, GroundingDINO—a pure detector with no conversational capabilities—can now successfully ground references it could not handle in the baseline condition.

\paragraph{Rewrite Quality Validation.}
 Validation of rewrites shows 83.4\% correctly preserve referent identity, with grounding success strongly correlating with rewrite quality: good rewrites achieve 55.0\% (IoU$\geq$0.5), borderline 32.5\%, bad 23.4\%— confirming genuine disambiguation rather than fortuitous errors. Error analysis reveals systematic patterns (wrong object: 7.5\%; missing part 
relations: 20.3\%; details in Appendix ~\ref{app:rewrite_validation}).

\subsubsection{Rewriting Results}

Table~\ref{tab:rewrite_results} presents results for our two-stage pipeline. The improvements are substantial and consistent across all models and reference types.

Pronominals show the most dramatic improvements. GroundingDINO jumps from approximately 5\% to 56.7\%, Florence-2 from 25.8\% to 48.9\%, and Qwen2.5-VL from 30.4\% to 50.3\%. The massive gain for GroundingDINO is particularly revealing: this model has no ability to process conversational context, yet after rewriting it outperforms Qwen2.5-VL's baseline with context (56.7\% vs 30.4\%). This demonstrates that the bottleneck is not visual grounding capability, but rather the conflation of linguistic and visual reasoning.

Partitive NPs also benefit substantially. GroundingDINO improves from approximately 15\% to 49.5\%, Florence-2 from 21.6\% to 39.7\%, and Qwen2.5-VL from 29.6\% to 40.8\%. The rewriter adds explicit part-whole relationships that models struggle to infer from conversation alone. GroundingDINO again achieves the highest absolute performance (49.5\%), nearly doubling Qwen's baseline.

Full NPs show varied patterns. Qwen2.5-VL shows minimal gain (+1.2\%) because it already handles explicit references well (53.2\% baseline). However, Florence-2 and GroundingDINO benefit substantially from rewriting, which removes object IDs, preserves qualifiers, and adds spatial context.

Remarkably, GroundingDINO with rewriting achieves the highest performance across all categories (61.1\% / 49.5\% / 56.7\%), outperforming Qwen2.5-VL's best baseline with context (53.2\% / 29.6\% / 30.4\%). This is despite GroundingDINO having no native ability to process conversational history. Mean IoU remains high (77--89\%) across all models, indicating that when models successfully identify the correct object, they localize it precisely. The challenge is identification, not localization—which our rewriting approach directly addresses.

\section{Discussion and conclusions}

In this work, we present a multimodal benchmark for situated referential communication, combining spontaneous VR dialogue with synchronized 3D scene data. This enables analysis of grounding in realistic, multimodal contexts beyond prior datasets.

A human text-only evaluation established a lower bound for grounding performance. Participants easily resolved explicit noun phrases but relied on dialogue history for ambiguous partitive and pronominal references. By removing non-verbal cues, this setup provides a fair baseline for comparison with current vision–language models (VLMs).

Compared to this baseline, state-of-the-art VLMs such as Florence-2 and GroundingGPT exhibit a clear domain gap: trained on self-contained captions, they struggle with the spontaneous, ambiguous language dominant in our data. With nearly 50\% of references being pronominal, this failure affects the majority of situated expressions. The gap between human text-only and full multimodal performance reveals a larger \textit{multimodal gap}, the benefit humans gain from gaze and gesture cues, highlighting our dataset’s unique potential to close it.

Our two-stage pipeline, which separates linguistic reasoning from visual localization, outperforms monolithic VLMs and shows that resolving ambiguity prior to grounding yields more robust results. Analysis further shows that:
\begin{itemize}
\item Conversational history benefits both humans and models, especially for ambiguous references.
\item Added context can slightly lower match rates but increase localization precision (IoU), indicating a trade-off between recall and spatial accuracy.
\item Even when semantically off, VLM predictions remain spatially close to ground truth.
\end{itemize}

While effective, our  model introduces latency. A practical next step is teacher–student distillation, where GPT-4o converts ambiguous expressions into explicit, grounded forms to train faster, deployable models.

Our current 2D text–image evaluation forms only the first layer of analysis. Future work will exploit synchronized temporal and non-verbal streams to enable:
(1) video-based grounding over unfolding interactions, 
(2) gaze-conditioned attention mechanisms, and 
(3) full 3D scene reasoning for embodied references.

Ultimately, progress in multimodal grounding requires moving beyond static image–text alignment toward models capable of temporal reasoning, dialogue-aware interpretation, and multimodal integration. By providing a rigorous baseline and identifying key gaps, our benchmark lays the groundwork for the next generation of situated AI systems. The full dataset, with synchronized motion, speech, and gaze, will be publicly released to support advances in both grounding and embodied behavior modeling.

\section{Bibliographical References}\label{sec:reference}

\bibliographystyle{lrec2026-natbib}
\bibliography{lrec2026-example}

\section{Language Resource References}
\label{lr:ref}
\label{sec:language-resources}

\subsection{Resource to be Shared}
We release \textbf{MM-Conv}, a multimodal benchmark for context-aware grounding in situated 3D dialogue. The resource comprises synchronized \emph{speech}, \emph{text}, \emph{egocentric RGB}, \emph{metric depth (normalized for the starter pack)}, \emph{segmentation masks}, \emph{full-body motion}, \emph{gaze}, \emph{facial blendshapes}, and \emph{3D scene geometry}.

\paragraph{Scope and Scale.}
\begin{itemize}
    \item Duration: \textasciitilde6.7 hours of dyadic interaction (VR).
    \item Instances: 4{,}211 referring expressions (after filtering: 4{,}001 visible).
    \item Linguistic forms: full NP, partitive/attribute NP, pronominal.
    \item Speech: \textasciitilde250k transcribed words with word-level timecodes.
    \item Visual: Five AI2-THOR apartment rooms, egocentric frames with per-pixel masks.
\end{itemize}

\paragraph{Modality Summary.}
\begin{table}[h]
\centering
\small
\begin{tabular}{ll}
\toprule
\textbf{Modality} & \textbf{Details} \\
\midrule
Speech/Text & WhisperX transcripts+word timings \\
RGB/Depth & Egocentric images+normalized depth\\
Segmentation & Per-pixel object IDs and GT masks \\
Motion & Full-body (main+interlocutor) \\
Hands/Face & Finger tracking, 52 facial blendshapes \\
Gaze & Binocular gaze from HMD \\
Scene Graph & Object instances and 3D metadata \\
\bottomrule
\end{tabular}
\caption{Modalities included in the resource.}
\label{tab:lr-modalities}
\end{table}

\paragraph{Format.}
Each RE is aligned to its egocentric frame; we provide JSON annotations (utterance, token indices, category, referent ID), RGB, mask, depth (normalized in the sample pack), and metadata. 

\paragraph{Availability and License.}
The resource will be made available for research use upon publication under \textbf{CC BY-NC 4.0}. A \emph{starter pack} (\(\leq\)20\,MB) with schemas, 10--20 samples, and an evaluation script is provided for review. The full release (with documentation and datasheet) will follow camera-ready.

\paragraph{Ethics and Privacy.}
Data were collected under informed consent; audio is transcribed and anonymized; egocentric renders are from simulation (no real faces). We remove metadata that could reveal identity or location. Redistribution is limited to non-commercial research; re-identification is prohibited.


\subsection{Resources Used in This Work}
We used the following LRs/tools; each is (or will be) separately entered in the LRE Map:

\begin{itemize}
    \item \textbf{AI2-THOR} simulator for interactive 3D environments~\citetlanguageresource{AI2THOR}.
    \item \textbf{WhisperX} for ASR with word-level alignment~\citeplanguageresource{WhisperX}.
    \item \textbf{Florence-2} (tool)~\citeplanguageresource{Florence2}; paper~\cite{xiao2024florence}.
    \item \textbf{GroundingGPT} (tool)~\citeplanguageresource{GroundingGPT}; paper~\cite{li2024groundinggpt}.
    \item \textbf{OptiTrack} motion capture system~\citeplanguageresource{OptiTrack}.
    \item \textbf{MANUS Quantum MetaGloves} for finger tracking~\citeplanguageresource{MANUS}.
\end{itemize}

\subsection{Reproducibility and Evaluation Artifacts}
We provide a minimal evaluation script (IoU, Match@\{0.3,0.5\}) and JSON schemas for predictions/ground truth to standardize reporting. Benchmarks include single-object full/partitive/pronominal subsets with and without dialogue context.

\newpage
\appendix
\section{Data Collection and Experimental Setup}
\label{appendix:collection}
This section details the methodology for collecting our multimodal dataset. We combine motion capture with virtual reality, recording conversations within the AI2-THOR physics simulator\footnote{\url{https://github.com/allenai/ai2thor}}. This setup provides experimental control for replicating conditions precisely and simplifies the annotation of objects and scenes.

\subsection{Experimental Scenarios}
To elicit natural, spontaneous referential language, participants engaged in one of three conversational scenarios within the virtual environment. The setup involved a main actor (the "speaker," wearing the VR headset) and a secondary actor (the "interlocutor"). The scenarios were designed to encourage descriptive language and interaction centered on objects in the scene.

\begin{itemize}
    \item \textbf{Scenario 1: Bragging/Introducing New Apartment}
        \begin{itemize}
            \item \textit{Roles:} Main Actor: Apartment Owner, Secondary Actor: Friend.
            \item \textit{Description:} The apartment owner shows off their new apartment to their friend, highlighting various features and objects in the room.
            \item \textit{Dialogue Focus:} Description of the objects, personal anecdotes about their acquisition, and the benefits of each item.
        \end{itemize}

    \item \textbf{Scenario 2: Landlord Asking About Objects}
        \begin{itemize}
            \item \textit{Roles:} Main Actor: Landlord, Secondary Actor: Tenant.
            \item \textit{Description:} The landlord inquires about various objects in the apartment, possibly checking for maintenance needs or understanding the tenant's living conditions.
            \item \textit{Dialogue Focus:} Questions about the objects, their usage, condition, and any issues.
        \end{itemize}

    \item \textbf{Scenario 3: Interior Designer Giving Tips}
        \begin{itemize}
            \item \textit{Roles:} Main Actor: Interior Designer, Secondary Actor: Client.
            \item \textit{Description:} The interior designer provides suggestions and advice on improving the apartment's aesthetics and functionality.
            \item \textit{Dialogue Focus:} Suggestions for rearranging furniture, adding new decor items, and making the space more efficient.
        \end{itemize}
\end{itemize}

\subsection{Scene Details and Object Distribution}
\begin{figure}[htb]
  \centering
  \includegraphics[width=0.95\linewidth]{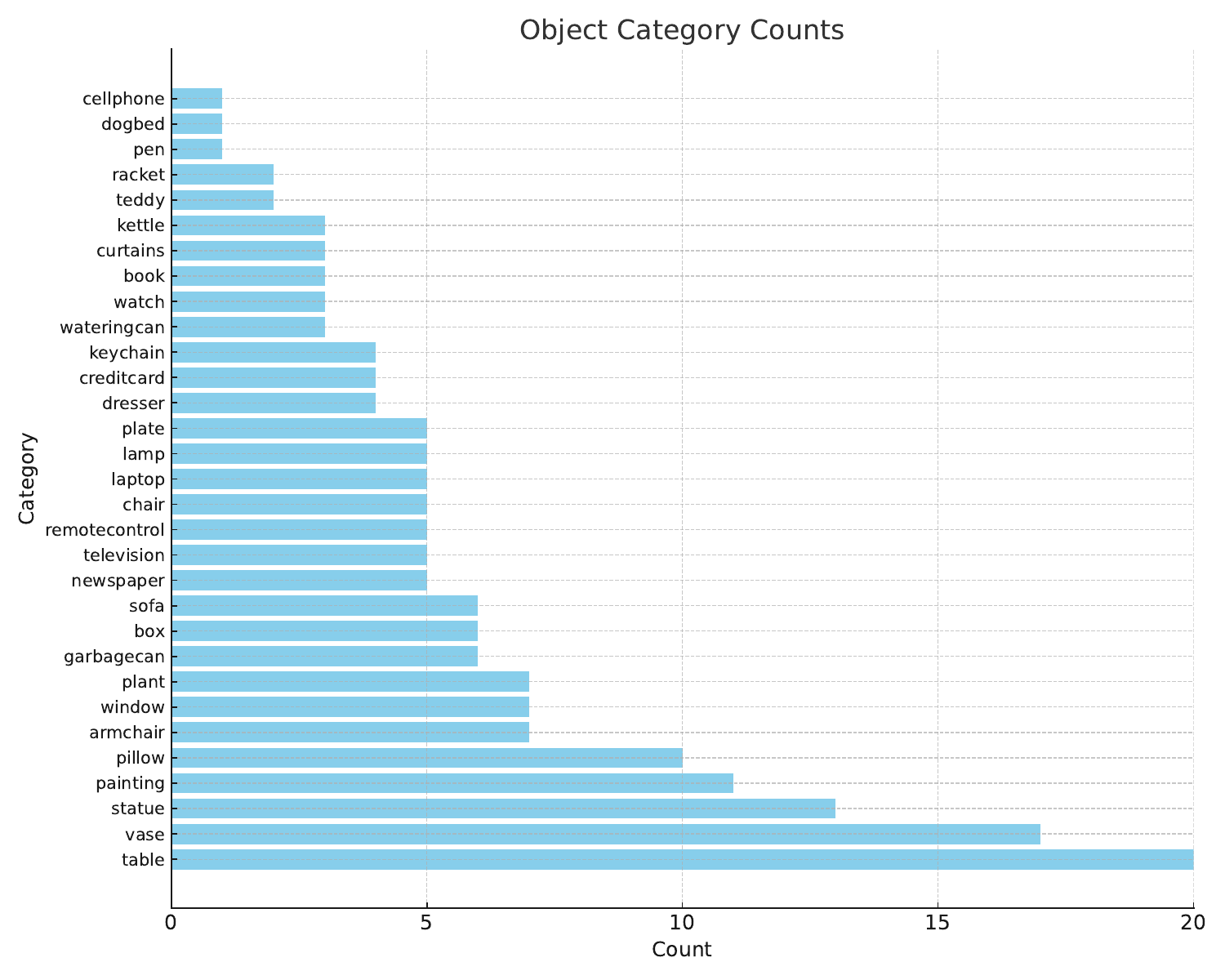}
  \caption{Object category distribution across simulated environments used in the experiments.
  }
  \label{fig:object_dist}

\end{figure}

The experiments were conducted across five different virtual rooms from the AI2-THOR simulator. The average number of interactable objects in these rooms was $38 \pm 3.16$. This variety ensures a diverse distribution of objects and spatial layouts. Figure \ref{fig:object_dist} shows the frequency of different object categories across all environments used in the study.

\subsection{Hardware and Software Setup}
\begin{figure}[htb]
  \centering
  \includegraphics[width=0.85\linewidth]{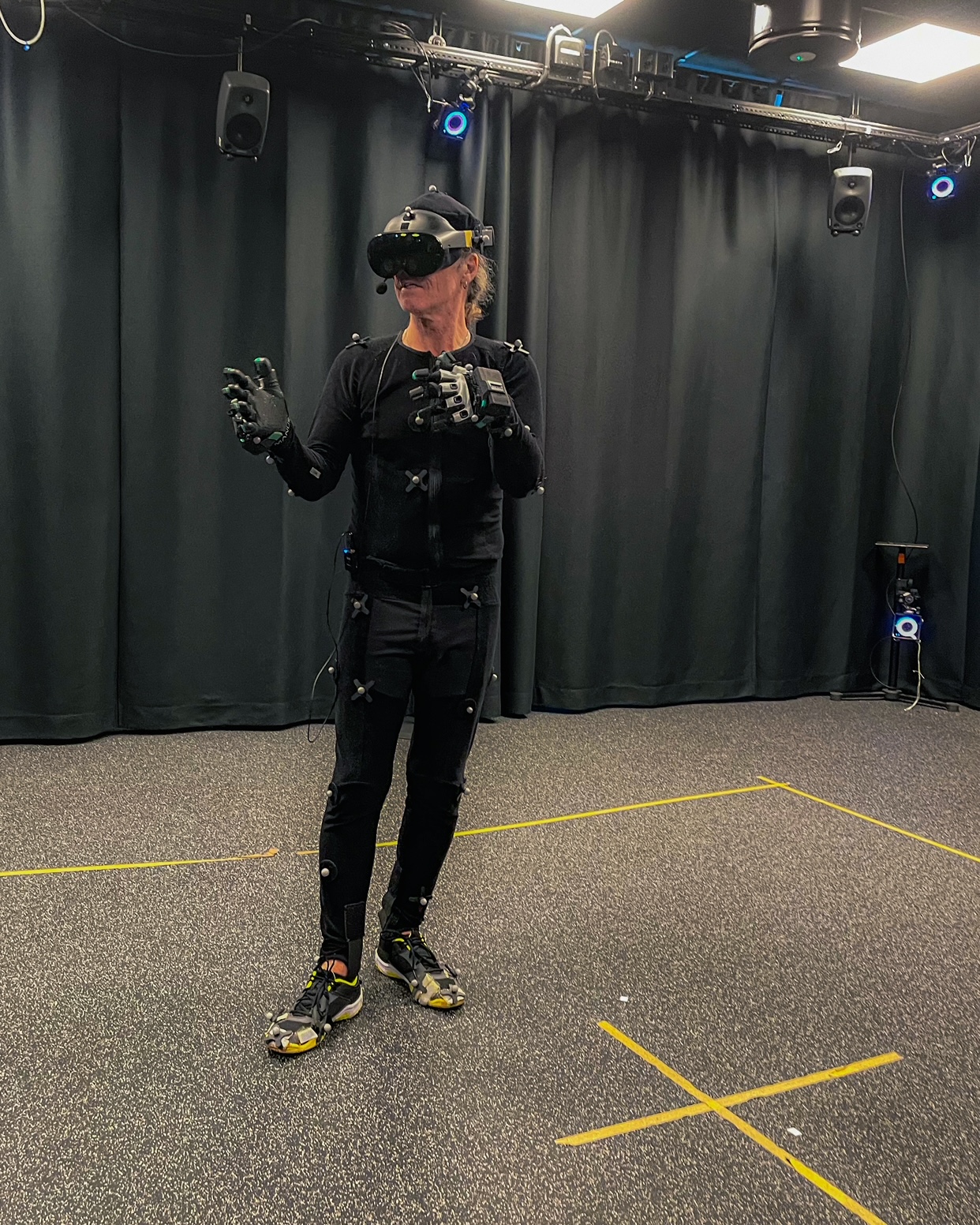}
  \caption{Illustration of hardware setup in motion capture lab.
  }
\label{fig:hardware_setup}

\end{figure}

Our data capture system integrated several components to record synchronized multimodal data streams. An illustration of the hardware setup is provided in Figure \ref{fig:hardware_setup}.

\begin{itemize}
    \item \textbf{Motion Capture:} Skeletal information was recorded using an OptiTrack system\footnote{\url{https://optitrack.com/}} with 16 Prime x 41 cameras, tracking 50-marker skeletons for both participants.
    \item \textbf{Finger Tracking:} Accurate finger movements were captured using Quantum Mocap Metagloves\footnote{\url{https://www.manus-meta.com/products/quantum-mocap-metagloves}}.
    \item \textbf{VR and Gaze Tracking:} A META Quest Pro headset\footnote{\url{https://www.meta.com/quest/quest-pro/}} was used to immerse the main speaker in the virtual environment and to record face and gaze tracking data.
    \item \textbf{Simulation Environment:} The virtual scenes were rendered in the AI2-THOR physics simulator\footnote{\url{https://github.com/allenai/ai2thor}}.
    \item \textbf{Synchronization:} Tentacle Sync E devices\footnote{\url{https://tentaclesync.com/sync-e}} were used to generate SMPTE time codes, ensuring fine-grained synchronization between the audio recordings, OptiTrack motion data, and AI2-THOR simulation data. The position of the Quest headset was aligned with the motion-captured head position to ensure visual and physical synchrony.
\end{itemize}

\section{Annotation Pipeline Details}
This section details the methodology for annotating our multimodal dataset. 
\subsection{Speech Transcription and Alignment}
\label{appendix:labelstudio}

To support speech annotation, we used Label Studio\footnote{\url{https://labelstud.io}} to verify and correct the generated transcriptions. Each annotation consisted of a pair of linked elements: a timed audio chunk and its corresponding transcription text box. This interface allowed annotators to listen to speech segments in context and make corrections directly on the aligned text.

We initialized Label Studio using audio chunk boundaries provided by WhisperX's voice activity detection (VAD) module, yielding more accurate and robust segmentation compared to sentence-level heuristics. Each chunk was pre-filled with WhisperX’s ASR output, providing a fast starting point for correction. Annotators focused on preserving spontaneous speech characteristics (e.g., filled pauses, repetitions) while ensuring accuracy and clarity.

To assist with triage and quality control, we implemented color-coding in the annotation interface: segments containing non-speech events—such as laughter or spoken noise—were visually highlighted in distinct colors. This allowed annotators to quickly identify and skip or deprioritize these segments during transcription.

As only one of the speakers is a native speaker, annotators also took care to check for effects of accented speech, such as the Spanish native's pronunciation of ``vase'' being recognized as ``base''.

\begin{figure}[h]
  \centering
  \includegraphics[width=\linewidth]{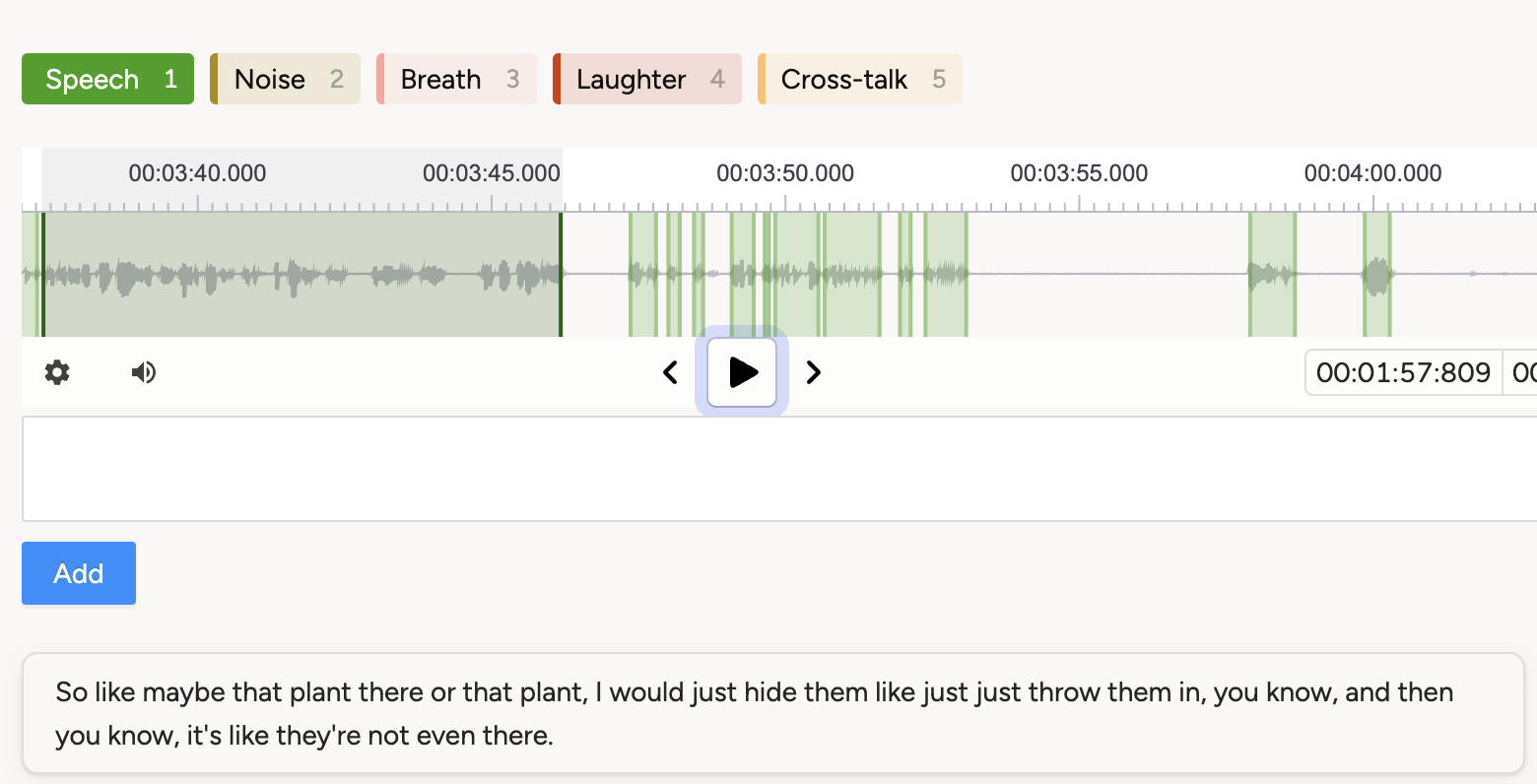}
  \caption{Label Studio interface for speech annotation. Annotators reviewed WhisperX transcriptions aligned to VAD-based audio segments and edited text within the linked text box.}
  \label{fig:labelstudio}
\end{figure}

\newpage
\subsection{GPT-4o prompts}

\subsubsection{Reference Classification Prompt}
The following prompt was used to classify referring expressions into semantic categories using GPT-4o:

\begin{center}
\begin{minipage}{0.5\textwidth}
\begin{lstlisting}[
    caption={Prompt used for classifying referring expressions into semantic categories using GPT-4o.},
    label={lst:ref-classification},
    basicstyle=\ttfamily\small,
    breaklines=true,
    breakatwhitespace=false,
    frame=single
]
You are analyzing a sentence and a highlighted phrase within it. 
The object of interest is called: '{object_name}'.

Utterance: '{utterance}'  
Referring phrase: '{phrase}'

Your task is to categorize the referring phrase into one of the following types:
- 'exact': matches or semantically refers to the full object name 
  (e.g., 'lamp' -> 'lamp', or 'couch' -> 'sofa')
- 'part': refers to a part, feature, or something depicted within the object 
  (e.g., 'handle' of a 'pan', 'cloud' in a 'painting', or 'the text' on a 'poster')
- 'pronominal': is a pronoun (e.g., 'this', 'it', 'that one', 'they')

Important note:  
- If the phrase refers to something inside or depicted on a painting, poster, or picture, classify it as 'part'.  
- For example, 'the dog' referring to an image in a painting -> label: 'part'  
- But if the phrase is 'the painting', 'the poster', or 'the picture' -> label: 'exact'

Only return the label: 'exact', 'part', or 'pronominal'.
\end{lstlisting}
\end{minipage}
\end{center}

\subsubsection{Contextual Disambiguation Prompt}
We use Qwen2.5-VL for linguistic disambiguation in Stage~1 of the
two-stage pipeline (Section~4.3.2).  The rewriter operates in
\emph{text-only} mode: no image is provided, so disambiguation is
pure linguistic reasoning over dialogue history and the visible-object
list.  The rewritten noun phrase is then passed to the Stage~2 visual
grounding model (Florence-2, GroundingDINO, or Qwen2.5-VL).

\begin{center}
\begin{minipage}{0.5\textwidth}
\begin{lstlisting}[
    caption={Prompt used for contextual disambiguation
             of referring expressions using Qwen2.5-VL.
             The model receives conversation history and
             a visible-object list, and outputs a single
             noun phrase.},
    label={lst:ref-disambiguation},
    basicstyle=\ttfamily\small,
    breaklines=true,
    breakatwhitespace=false,
    frame=single
]
SYSTEM: You are a referring-expression rewriter for visual grounding. Given a phrase, its utterance, conversation history, and a list of visible objects, output a single unambiguous noun phrase (3-8 words) that identifies the intended referent.

TASK: Rewrite the phrase "{phrase}" from the utterance into a clear, specific noun phrase that can be grounded in the visible scene.

UTTERANCE: {utterance}

CONVERSATION HISTORY:
{conversation_history}

VISIBLE OBJECTS IN SCENE:
{object_list}

CRITICAL RULES:
1. OUTPUT ONLY A NOUN PHRASE - no explanations, no meta-commentary
2. If the original phrase doesn't match  visible objects exactly, output the CLOSEST semantic match
3. Add spatial/relational context when helpful (e.g., "the desk with the laptop" vs just "the desk")
4. Use object attributes from the  visible objects list when available
5. NEVER output: "There is no X   visible", "The scene does not  contain", "I cannot see"
6. NEVER use quotation marks, backticks,
   or formatting in your output
7. Keep it concise but specific
   (typically 3-8 words)

EXAMPLES:
- "the stove" + visible: [FirePlace] -> "the fireplace"
- "it" + context about painting  + visible: [Painting2] -> "the wall painting"
- "the little black thing"  + context: helicopter painting  -> "the black element in the painting"
- "cloud" + visible: [Painting_0db0fb84]  -> "the cloud-like shape in the  painting"

YOUR REWRITE (noun phrase only):
\end{lstlisting}
\end{minipage}
\end{center}

\section{Rewrite Quality Validation}
\label{app:rewrite_validation}

Validation of rewrites shows 83.4\% correctly preserve referent identity.  Performance varies by type: pronominals 87.2\%, full NPs 84.6\%, partitives 67.4\%.

\begin{table}[h!]
\centering
\caption{Grounding success by rewrite quality.}
\label{tab:grounding_by_quality_full}

\begin{tabularx}{\linewidth}{@{}lrrr@{}}
\toprule
\textbf{Quality}  & \textbf{Acc@0.5} & \textbf{Acc@0.3} & \textbf{Mean IoU} \\
\midrule
Good        & 55.0\% & 64.7\% & 0.481 \\
Borderline  & 32.5\% & 44.6\% & 0.323 \\
Bad         & 23.4\% & 27.2\% & 0.216 \\
\midrule
\textbf{Overall} & \textbf{50.5\%} & \textbf{60.0\%} & \textbf{0.447} \\
\bottomrule
\end{tabularx}

\vspace{2pt}
\footnotesize
\begin{tabularx}{\linewidth}{@{}X@{}}

\textbf{Error-flag impact} (Acc@0.5): wrong\_object 7.5\% | missing\_part 20.3\% | 
lost\_qualifier 41.8\%. Cross-rewriter: GPT-4o 51.9\% vs Qwen 55.0\% (3.1pp).
\end{tabularx}

\end{table}

\end{document}